\documentclass[sigconf]{acmart}
\AtBeginDocument{%
  }

\setcopyright{rightsretained}
\copyrightyear{2025}
\acmYear{2025}
\acmConference{SA Posters '25}{December 15-18, 2025}{Hong Kong, Hong Kong}
\acmBooktitle{SIGGRAPH Asia 2025 Posters (SA Posters '25), December 15-18, 2025}
\acmDOI{10.1145/3757374.3771423}
\acmISBN{979-8-4007-2134-2/2025/12}

\acmSubmissionID{pos_143}

\usepackage{colortbl} 

\begin{document}

\title{Foveation Improves Payload Capacity in Steganography}

\settopmatter{authorsperrow=4}

\author{Lifeng Qiu Lin}
\orcid{0009-0005-4720-665X}
\affiliation{%
  \institution{University College London}
  \city{London}
  \country{UK}
}
\email{lifeng.qiu.lin.lq@gmail.com}

\author{Henry Kam}
\orcid{0009-0002-5913-4647}
\affiliation{%
\institution{New York University}
\state{New York}
\country{USA}}
\email{henry.j.kam@gmail.com}

\author{Qi Sun}
\orcid{0000-0002-3094-5844}
\affiliation{%
\institution{New York University}
\state{New York}
\country{USA}
}
\email{qisun0@gmail.com}

\author{Kaan Akşit}
\orcid{0000-0002-5934-5500}
\affiliation{%
  \institution{University College London}
  \city{London}
  \country{UK}
}
\email{kaanaksit@kaanaksit.com}

\renewcommand{\shortauthors}{Qiu et al.}

\begin{abstract}
Steganography finds its use in visual medium such as providing metadata and watermarking.
With support of efficient latent representations and foveated rendering, we trained models that improve existing capacity limits from 100 to 500 bits, while achieving better accuracy of up to 1 failure bit out of 2000, at 200K test bits.
Finally, we achieve a comparable visual quality of 31.47 dB PSNR and 0.13 LPIPS, showing the effectiveness of novel perceptual design in creating multi-modal latent representations in steganography.

\end{abstract}



\maketitle

\section{Introduction}

Steganography concerns about hiding data in another medium \cite{wang2023data}. More specifically, our work studies embedding information in images.
Steganography is therefore useful in conveying multi-modal information such as labels, scene descriptions, or copyright marks. 
Growing number of AI-generated content and introduction of AR/VR systems increases its importance further by broadening the application scope \cite{rezaei2024lawa}.

\begin{figure}[h]
  \centering
  \includegraphics[width=\linewidth]{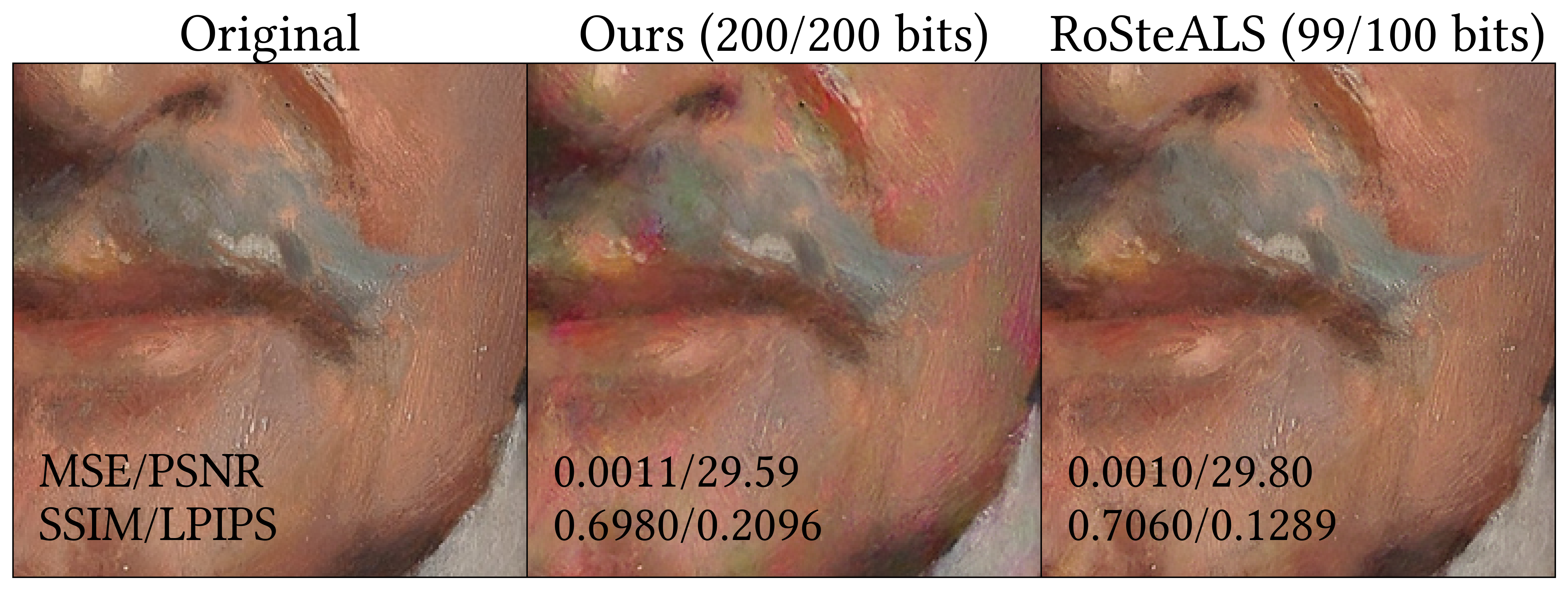}
  \caption{Visual example of stego and payload recovery using our proposed foveated steganography and RoSteALS (Source: MetFaces \cite{karras2020training}).}
  \Description{Three pictures from left to right: original, our produced stego, and RoSteALS produced stego, with ours higher capacity and accuracy.}
  \label{fig:vis}
\end{figure}

Our work leverages latent representations \cite{yilmaz2024learned} and a foveated rendering loss \cite{walton2022metameric} to increase payload capacity in steganography. 
With only 2000 training images, we achieve bit accuracy of 99.99\% for 40K test set bits.
Concerning state of the art in latent methods \cite{bui2023rosteals}, our approach increases the payload capacity from 100 bits up to 500 bits with up to 100\% recovery under non-distortion condition, 27.56 db PSNR and 0.26 LPIPS. 
Our final contribution is the introduction of Metameric Foveated Rendering loss in steganography, which noticeably improves all visual metrics and quality with respect to classic L2 loss.

\section{Method}

Considering a message payload $P \in \{0, 1\}^k$ consisting of $k$ bits and a input image (cover) $I \in \mathbb{R}^{h \times w \times c}$, find two functions $H$ and $R$, to produce an output image (stego), $H(I, P) = I' \in \mathbb{R}^{h \times w \times c}$, and output payload, $R(I') = P' \in \{0, 1\}^k$. The aim is to reduce the distortion between $I$ and $I'$ while maximizing the accuracy between $P$ and $P'$.

\begin{figure}[h]
  \centering
  \includegraphics[width=0.8\linewidth]{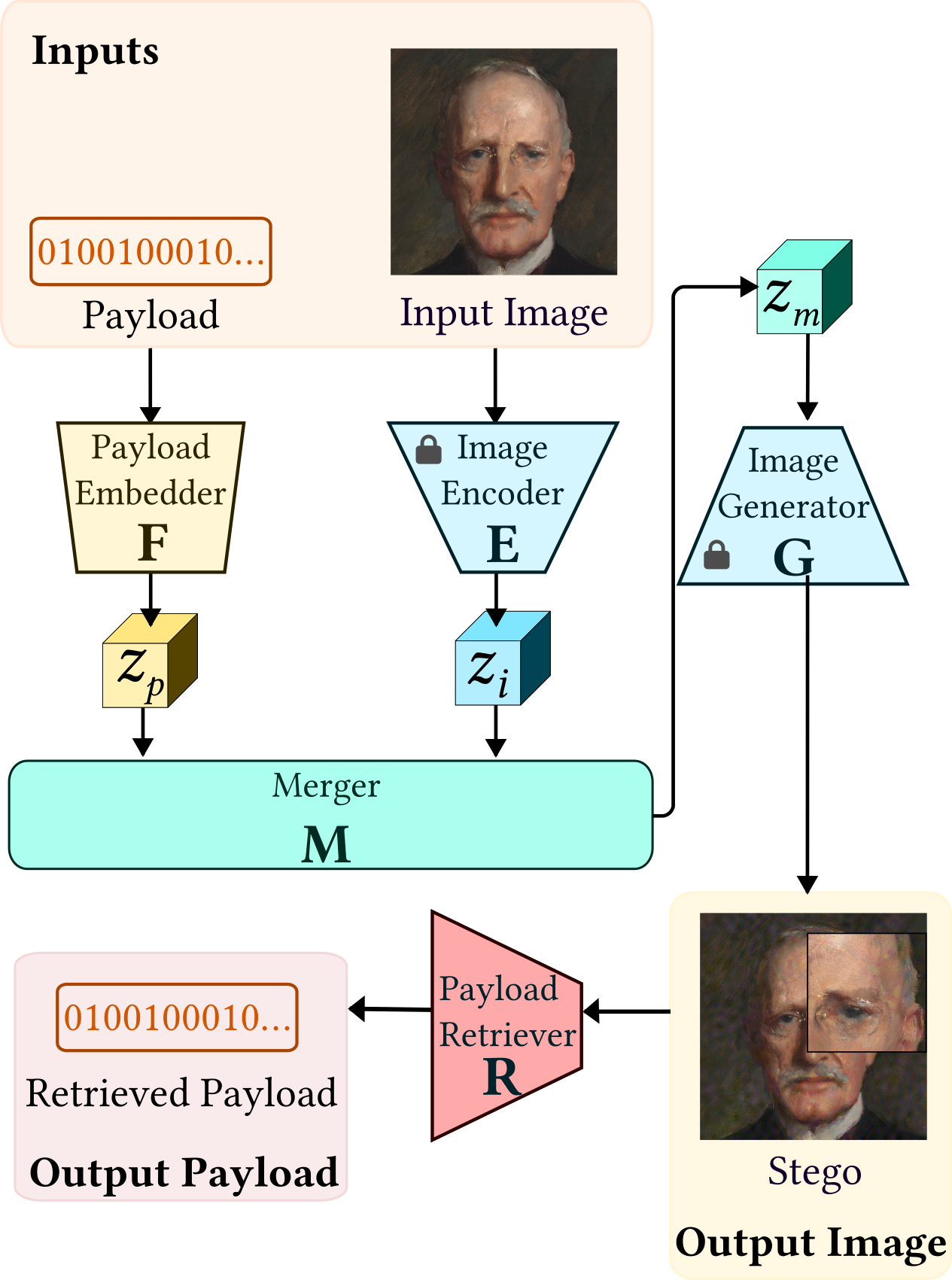}
  \caption{Our proposed foveated steganography approach (Source: MetFaces \cite{karras2020training}).}
  \Description{A diagram showing the architecture of the proposed system. Inputs are payload and input image, they go through payload embedder and image encoder respectively, producing payload and image latent representations. Merger combines these two latent representations and let image generator reconstruct the output stego image. Finally, a payload retriever extracts the output payload from the output image.}
  \label{fig:architecture}
\end{figure}

\begin{table*}
  \caption{Table of results. Baseline is trained on MSE. RoSteALS uses MSE \& LPIPS. Metameric solely relies on Metameric Loss.}
  \label{tab:results}
  \begin{tabular}{lcccccccccc}
    \toprule
    Experiment & Resolution & Capacity & Bit Accuracy & MSE & PSNR & SSIM & LPIPS & Metameric Loss\\
    \midrule
    Benchmark Baseline (RoSteALS) & 256 & \cellcolor{red!25}100 & 0.9942 & \cellcolor{green!25}0.0009 & \cellcolor{green!25}32.16 & \cellcolor{yellow!25}0.8971 & \cellcolor{green!25}{0.0780} & \cellcolor{green!25}0.0016 \\
    Vanilla Baseline & 256 & \cellcolor{red!25}100 & \cellcolor{green!25}{1} & {0.0015} & 28.90 & \cellcolor{yellow!25}0.8833 & 0.1621 & 0.0072 \\
    Vanilla Baseline & 256 & \cellcolor{yellow!25}200 & \cellcolor{green!25}{1} & 0.0018 & 28.37 & 0.8681 & 0.2047 & 0.0072 \\
    Metameric Baseline & 256 & \cellcolor{yellow!25}200 & 0.9998 & \cellcolor{green!25}{0.0010} & \cellcolor{yellow!25}{31.47} & \cellcolor{green!25}0.8871 & {0.1288} & \cellcolor{green!25}{0.0017} \\
    Vanilla Baseline & 256 & \cellcolor{green!25}{500} & 0.9997 & 0.0021 & 27.56 & 0.8348 & 0.2613 & 0.0077 \\
    Metameric Baseline & 256 & \cellcolor{green!25}500 & 0.9998 & 0.0013 & 30.08 & 0.8617 & 0.1841 & 0.0024 \\
    Vanilla Baseline & 128 & \cellcolor{green!25}{500} & \cellcolor{red!25}{0.4997} & 0.0007 & 33.11 & 0.9061 & 0.0570 & 0.0023 \\
    \bottomrule
  \end{tabular}
\end{table*}

Our framework approaches this problem as depicted in Figure~\ref{fig:architecture}. 
In the hiding stage, a frozen image encoder, $E$, transforms input image into a latent representation, $E(I) = Z_i$. 
Payload embedder, $F$, creates also a learned representation, $P(I) = Z_p$. 
Together, they are manipulated by the merger, $M$, producing a merged latent, $M(Z_i, Z_p) = Z_m$, which a frozen image generator, $G$, uses to reconstruct the output image,  $G(Z_m) = I'$. 
Finally, a payload retriever, $R$, extracts the output payload, $R(I') = P'$.
The loss function is defined as the combination of payload and image quality losses, being BCE and Metameric Foveated Rendering \cite{walton2022metameric} (defaulted to center) losses respectively.
Formally,
\begin{math}
  \mathcal{L}_{total} = \mathcal{L}_{payload} + \lambda_i \cdot \mathcal{L}_{image} \nonumber = \operatorname{BCE}(P, P') + \lambda_i \cdot (\operatorname{MetamericLoss}(I, I')),
  \label{eq:loss}
\end{math}
where $\lambda_i$ controls the trade-off between the two losses.

The dataset is a balanced mixture of 2000 training, 400 validation, and 400 test images from MetFaces \cite{karras2020training} and CLIC datasets \cite{toderici2020workshop}. 
For preprocessing, images are randomly cropped and padded to the size of input and normalized as autoencoder requires. 
Notably, this dataset is much smaller than typical datasets used for the same purpose, but is found sufficient to learn performing 100-bits steganography, within controlled computing resources, about two hours on a single RTX 4090 GPU.

\section{Results and Discussion}

The frozen pair of image encoder and image generator to create a high-quality latent representation, is the F4-with-attention version autoencoder from LDM VQGAN series \cite{rombach2022high}.
After evaluating empirically, we found its high reconstruction quality is suitable for the embedding process. Compared to other backbones, this one converges slower at payload embedding, but achieves better image quality in the end.
Keeping payload embedder as a fully connected layers is sufficient to encode the information after experimentation. 
For merger, the best performing architecture is adding two convolutional layer sandwiching the sum of image and payload latent, to soften the transition. 
Finally, ResNet50 was used as payload retriever as a popular and well-studied architecture. 

Apart from common metrics, we also report Metameric Loss, which is a perceptual criterion akin to foveated gaze.
Modeling the human visual system, this loss is more forgiving of visual distortions in the periphery and more harsh in the fovea.

The main results are shown in Table~\ref{tab:results}. 
Baselines achieved expanding payload capacities, at various resolutions.
At minimal setting, baseline has a bit accuracy of 99.99\%, failing to decode only 4 out of 40K test bits. 
Noticeably, we achieve 100\% recovery in the native resolution of benchmark, RoSteALS \cite{bui2023rosteals}, while other settings also all exceed 99.95\% compared to benchmark failing to reach 99.5\%.
Nevertheless, RoSteALS has better perceptual image quality. 
This is reasonable since it is an augmented version of the baseline which uses larger datasets, incorporates LPIPS in loss, and applies finer-grained optimization in the training.

Compared to baseline, Metameric Loss consistently improves the quality of the reconstructed images while keeping same level of bit accuracy. 
Compared to the benchmark, Figure~\ref{fig:vis} shows an example of resulting stego and recovered payload.
This shows effectiveness of this visual technique in enhancing perceptual fidelity of images.

Despite successfully unlocking higher message length, we notice tangible limits of payload capacity, such as failing to learn 500-bit payload at 128 resolution.
Resolution bounds the upper payload capacity under similar perceptual fidelity of images, and we hope to enhance this by introducing gaze as a new parameter.
Future directions include exploring robustness under various distortions, subjective experiments to compare visual quality, and ablation studies with same capacity benchmarks.
This work provides a light-weighted, human-centered, latent-based steganography framework which boosts payload capacity and accuracy while maintaining image quality.
By satiating the need of large capacity in transmitting messages, we step towards practical applications of steganography in real-world scenarios.


\bibliographystyle{ACM-Reference-Format}
\bibliography{refs}





\end{document}